\documentclass{article}
\usepackage{spconf,amsmath,graphicx}

\usepackage{amsmath,amsfonts,amssymb}
\usepackage{graphicx}
\graphicspath{{figures/}}
\usepackage[colorlinks=true, allcolors=blue]{hyperref}
\usepackage{breakcites}
\usepackage{tabu}
\usepackage{floatrow}
\usepackage{wrapfig}
\usepackage{booktabs}
\usepackage{makecell}

\newcommand{\ra}[1]{\renewcommand{\arraystretch}{#1}}

\title{Improving CNN Training using Disentanglement \\
for Liver Lesion Classification in CT}

\name{Avi Ben-Cohen$^{\star}$ \qquad Roey Mechrez$^{\dagger}$ \qquad Noa Yedidia$^{\star}$ \qquad Hayit Greenspan$^{\star}$}

\address{$^{\star}$ Tel Aviv University, Faculty of Engineering, Department of Biomedical Engineering\\
    $^{\dagger}$Technion - Israel Institute of Technology}

\begin{document} 

\newcommand*{\ShowNotes}{}

\definecolor{darkred}{rgb}{0.7,0.11,0.11}
\definecolor{darkgreen}{rgb}{0.11,0.7,0.11}
\definecolor{cyan}{rgb}{0.7,0.0,0.7}
\definecolor{dblue}{rgb}{0.2,0.2,0.8}
\definecolor{maroon}{rgb}{0.76,.113,.28}
\definecolor{burntorange}{rgb}{0.81,.33,0}

\ifdefined\ShowNotes
  \newcommand{\colornote}[3]{{\color{#1}\bf{#2: #3}\normalfont}}
\else
  \newcommand{\colornote}[3]{}
\fi

\newcommand {\todo}[1]{\colornote{cyan}{TODO}{#1}}
\newcommand {\avi}[1]{\colornote{magenta}{ABC}{#1}}
\newcommand {\hayit}[1]{\colornote{blue}{HG}{#1}}
\newcommand {\roey}[1]{\colornote{red}{RM}{#1}}

\maketitle

\begin{abstract}

Training data is the key component in designing algorithms for medical image analysis and in many cases it is the main bottleneck in achieving good results. Recent progress in image generation has enabled the training of neural network based solutions using synthetic data. A key factor in the generation of new samples is controlling the important appearance features and potentially being able to generate a new sample of a specific class with different variants. In this work we suggest the synthesis of new data by mixing the class specified and unspecified representation of different factors in the training data. Our experiments on liver lesion classification in CT show an average improvement of 7.4\% in accuracy over the baseline training scheme. 
\end{abstract}



\section{Introduction}
\label{sec:intro}  
A neural network trained for image classification is considered to be a black-box model to the user, as one can not identify what are the factors that contribute to a specific classification. Access to such information can reveal new insights and assists in classical medical practice of observing images.

The concept of \textit{disentangling} in the context of deep representation refers to separation of factors in the representations~\cite{mathieu2016disentangling}. The key idea is to separate the representation into complementary code which is associated with distinctiveness, i.e. distinguish the \textit{specified} factors and the \textit{unspecified} factors of the data. For example, if we would like to distinguish between different faces, the shape might be an unspecified factor where the appearance would be a specified factor. The main  goal of disentanglement is to allow the learning process to represent the information as two complementary vectors without constraining the latent space. As a result, the specified and unspecified factors definition is not accessible.  Mathieu et al.~\cite{mathieu2016disentangling} suggested to solve it by adversarial training, where label information is removed from the unspecified part of the encoding. 
In Hadad et al.~\cite{hadad2017two}  a simpler two-step disentanglement method was suggested: an encoder is first trained to satisfy a classifier, then this encoder is used in order to synthesize a new image with the same classification features but with different unspecified factors. 

\begin{figure}[t]
\centering
\includegraphics[width=\textwidth]{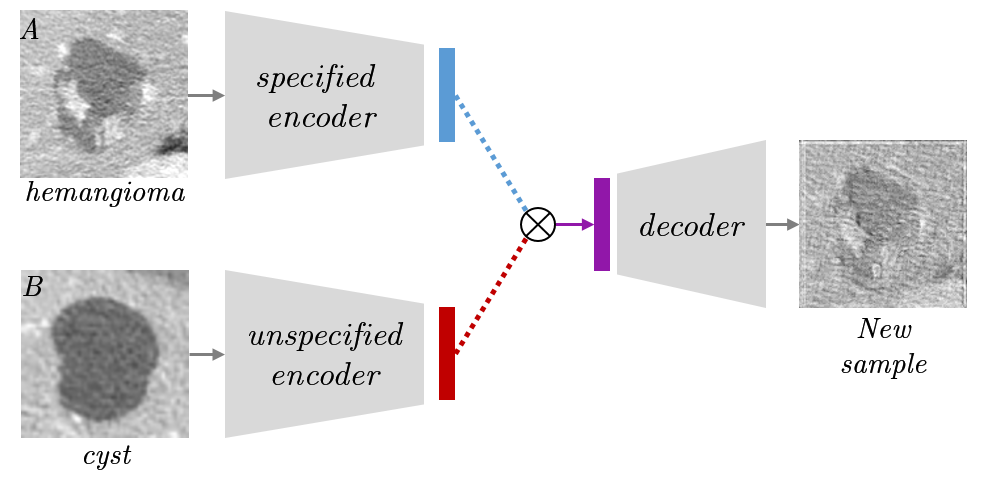}
\vspace{-1cm}
\caption{\textbf{The proposed generation-by-factor-mixture}: We combine the specified representation (blue) of sample $A$ with the unspecified representation (red) of sample $B$. The new representation is fed into the generator to result with a new sample. The disentangling training of the two encoders is the key factor which enables the mixture.}
\label{fig:concept}
\end{figure}

In this paper we follow \cite{hadad2017two} and use it as a pipeline for disentangling factors in the image representation of liver lesions. The separation is based on the relevance of a specific feature to the classification process of the image. For example, the contrast and texture can be significant factors for classification while the discriminative power of the general shape can be less important. We next use it in the training process by adding synthesized examples that include a mixture of the specified factors taken from different liver lesion types. Figure~\ref{fig:concept} shows an example of this concept by combining the unspecified representation of a cyst and mixing it with the specified representation of an hemangioma to create a new sample.

\begin{figure*}
\centering
\includegraphics[width=0.95\textwidth]{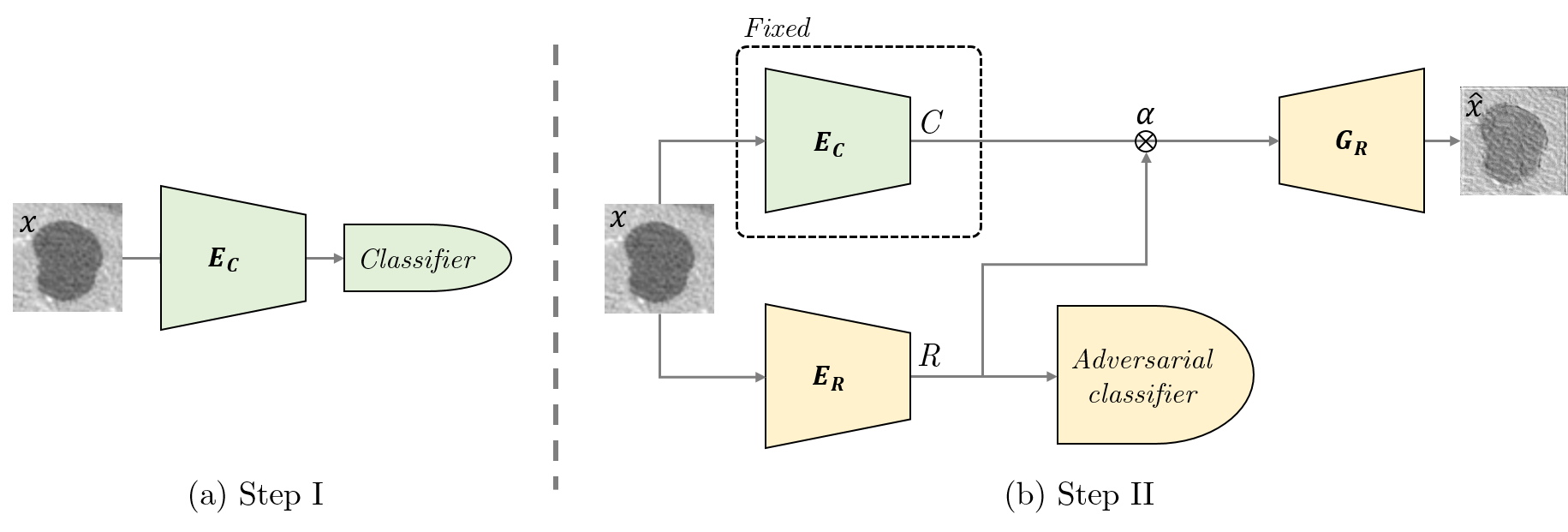}
\vspace{-0.5cm}
\caption{\textbf{The suggested method: }Our training process consists of two steps: in the first step (a) we train the encoder $E_C$ using a classification loss; in the second step (b) we train the encoder-decoder network $E_R, G_R$ using negative feedback from the classification loss and the reconstruction loss. At the bottleneck of the encoder-decoder we concatenate the information from the pre-trained encoder.
}
\label{fig:pipeline}
\end{figure*}

We focus our experiments on liver lesions classification in Computed tomography (CT) images. The liver can include multiple lesions from several types which makes it a complex task. Several steps  in automating the liver analysis can be found in recent literature: including liver segmentation, lesion detection, lesion categorization and follow-up. Among these,  the lesion {\it{segmentation}} task  has attracted the greatest deal of attention in recent years, being the focus of a couple of challenges. In practice, it is also important to separate between malignant and benign lesions. Frid-Adar et al.  \cite{frid2018gan} showed that generating synthetic medical data using generative adversarial networks \cite{goodfellow2014generative} can improve liver lesion classification results. In this study we present an alternative approach  to generate mixed samples within our dataset 
that challenge the training process and improve the classification system overall accuracy. 

The dataset used  includes 4 categories: cyst, metastasis, hemangioma and healthy liver parenchyma. After training our disentangling network, we used the synthetic samples in order to augment the training data. Our experiments show that  the proposed synthetic data augmentation improves the baseline results by up to $10\%$ in classification accuracy. 
\\
The contributions we present are two-fold:
\begin{enumerate}
\item To the best of our knowledge this is the first attempt to disentangle factors in medical images.  
\item We present a novel pipeline for synthesizing new data for training by swapping the specified and unspecified components of different liver lesion types. 
\end{enumerate}

\section{METHOD}
Our proposed method includes two major components: the first one is the disentanglement process 
and the second is the robust training process that makes use of the disentanglement to synthesize new examples with corresponding categorical vectors. In the following we elaborate on both components.

\subsection{Disentanglement of Liver Lesions}
Figure \ref{fig:pipeline} shows the general disentanglement pipeline used in this study. Similar to \cite{hadad2017two} our training process consists of two steps. In the first step an encoder $E_C$ is trained using the classic cross entropy classification loss. In the second step, we keep $E_C$ fixed and train an additional encoder termed $E_R$. 
$E_R$'s output is concatenated with $E_C$'s output. 
The concatenated vector, $z=(E_C(x),E_R(x))$ ,is then used as input to a decoder termed $G_R$ that is trained to reconstruct the original image using $L_2$ loss:
\begin{equation}
\mathcal{L}_{rec} = ||G_R(z)-x||_2^2
\label{eq:l1}
\end{equation}

In addition, an adversarial classifier is used to make sure the code derived from $E_R$ is not discriminative in terms of liver lesion classification. Hence, our overall loss function used in the second training step is as follows:
\begin{equation}
\mathcal{L}(E_R,G_R) = \mathcal{L}_{rec} + \lambda \cdot \mathcal{L}_{adv}
\label{eq:loss}
\end{equation}
where $\lambda$ is set to be negative in order to achieve the desired separation of the representation and $\mathcal{L}_{adv}$ is the cross entropy loss.

\subsection{Mixture of Specified Factors for Improved Classification}
\begin{figure} [h]
\setlength{\tabcolsep}{0.3em}
\centering
\begin{tabular}{cccc}
&$y_1$&$y_2$&$y_3$\\
&Hemangioma&Metastasis&Healthy\\
&
\includegraphics[width=0.23\textwidth]{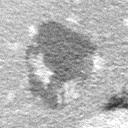}&
\includegraphics[width=0.23\textwidth]{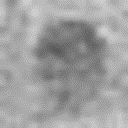}&
\includegraphics[width=0.23\textwidth]{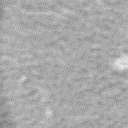}\\
\includegraphics[width=0.23\textwidth]{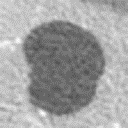}&
\includegraphics[width=0.23\textwidth]{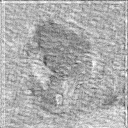}&
\includegraphics[width=0.23\textwidth]{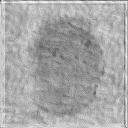}&
\includegraphics[width=0.23\textwidth]{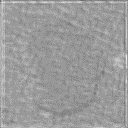}\\
$x$: Cyst  & (a) & (b) & (c)
\end{tabular}
\vspace{-0.3cm}
\caption{\textbf{Synthesizing new data}: Swapping the specified and unspecified components of a liver cyst cropped CT image. The images are generated using $R$ from the liver cyst image and $C$ from the top row images in the decoder. \textbf{(a)} $G_R(E_R(x), E_C(y_1))$ \textbf{(b)} $G_R(E_R(x), E_C(y_2))$ \textbf{(c)} $G_R(E_R(x), E_C(y_2))$.
}
\label{fig:swap}
\end{figure}

Using the encoders from the disentanglement process we can represent each image using the specified vector ($c$) and the unspecified vector ($r$). One possible manipulation would be to swap the vector $c$ of one image with the vector $c$ of another image. Figure~\ref{fig:swap} shows a swap example using the $r$ vector of a region of interest (ROI) of a cyst and swapping its $c$ vector with vectors that were extracted from different lesion types. Next, the $c$ and $r$ vectors are concatenated and $G_R$ is used to construct a new image. 

\begin{figure}[t]
\centering
\includegraphics[width=0.99\textwidth]{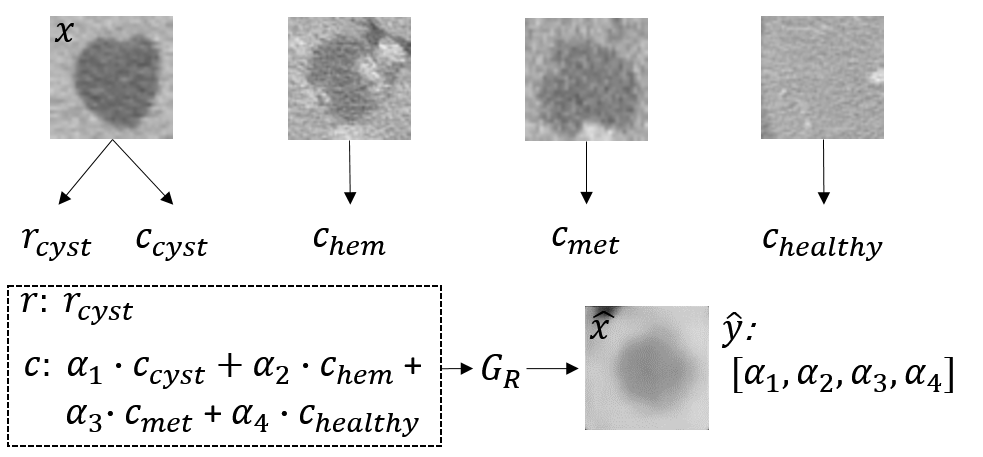}
\vspace{-0.3cm}
\caption{\textbf{Classification training scheme:} The specified vector of each sample is mixed with vectors from other classes to create new samples with new targets for the classification training.}
\label{fig:classification}
\end{figure}

Similar to the swapping example, we suggest a mixture of specified factors for training. For each image in the training process we extract its $r$ vector and create a new $c$ vector that includes a mixture of the original $c$ with other vectors from the different classes. The mixture is a sum of all vectors with different random proportion of each one. The different $c$ vectors are chosen as those that are most similar (in terms of $L_2$) to the original one to include a mixture $c$ vectors that are hard to distinguish but derive from different classes. The target classification vector is based on the proportion of each class in the mixture. Figure \ref{fig:classification} illustrates the proposed classification training scheme. Note that since the new generated samples are a mixture of different lesion classes we include the mixture in the target class probability vectors.

\subsection{Networks Architecture}
Three network architectures were used in this study: 1) for both encoders; 2) for the decoder; 3) for the classifier and the adversarial classifier. Table \ref{tab:architecture} provides a description of the models.

\begin{table}
\centering
	\ra{1.3}
	\begin{tabu}{@{}lllll@{}}
    \toprule
		\textbf{Encoder} & \phantom{ab} &  \textbf{Classifier} & \phantom{ab} & \textbf{Decoder}\\
    \midrule
C64K3S2   && Dense 256 && Dense 262144                                                   \\
C256K3S2 && BatchNorm  && \begin{tabular}[c]{@{}l@{}}Reshape\\ (16,16,1024)\end{tabular} \\
C1024K3S2 && Dense 256  && Dropout 0.5                                                    \\
Dense 256 && BatchNorm  && C1024K3 \\
         && Dense 4    && C256K3 \\
         && Softmax    && C64K3   \\
         &&            && C64K3   \\
         &&            && C1K3    \\
         &&            && Sigmoid \\
\bottomrule
\end{tabu}
\caption{\textbf{Network architecture} for the encoders, the classifiers, and the decoder. We used ReLU activation for all convolution layers and dense layers. We used the following notation: C64K3S2 is a Conv layer with 64 filters, kernel size $3\times3$ and stride 2. Each Conv layer in the decoder is followed by a x2 upsampling.}
\label{tab:architecture}
\end{table}

\section{Experiments and Results}
\label{sec:results}
\subsection{Dataset}
\begin{figure}[t]
\setlength{\tabcolsep}{.17em}
\centering
\begin{tabular}{cccc}
Cyst&Met.&Hem.&Healthy\\
\includegraphics[width=0.21\textwidth]{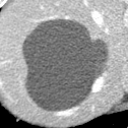}&
\includegraphics[width=0.21\textwidth]{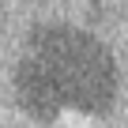}&
\includegraphics[width=0.21\textwidth]{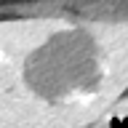}&
\includegraphics[width=0.21\textwidth]{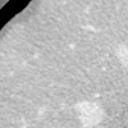}\\
\includegraphics[width=0.21\textwidth]{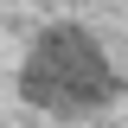}&
\includegraphics[width=0.21\textwidth]{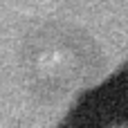}&
\includegraphics[width=0.21\textwidth]{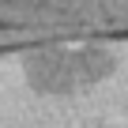}&
\includegraphics[width=0.21\textwidth]{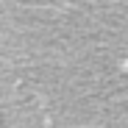}\\
\includegraphics[width=0.21\textwidth]{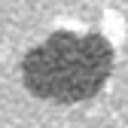}&
\includegraphics[width=0.21\textwidth]{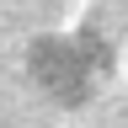}&
\includegraphics[width=0.21\textwidth]{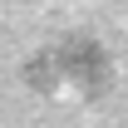}&
\includegraphics[width=0.21\textwidth]{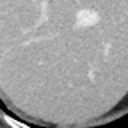}\\
\end{tabular}
\vspace{-0.4cm}
\caption{\emph{Real} data samples used for training including cyst, metastasis, hemangioma, and healthy liver parenchyma.}
\label{fig:data}
\end{figure}

The dataset used in this work contains cases of liver lesions collected from the Sheba Medical Center by searching medical records for cases of cysts, metastases and hemangiomas. Cases were acquired from 2009 to 2014 using four CT scanners: a General Electric (GE) Healthcare scanner, a Siemens Medical System scanner, a Philips Healthcare scanner and Toshiba Healthcare scanner, with the following parameters: $120kVp$, $120-630mAs$ and $0.9-5.0 mm$ slice thickness. Cases were collected with the approval of the institution’s Institutional Review Board. The input to our system are ROIs of lesions cropped from CT scans using the radiologist’s annotations. The ROIs are extracted to capture the lesion and its surrounding tissue relative to its size. Due to the large variability in lesion sizes, this results in varying size ROIs, afterwards all ROIs are resized to $128 \times 128$.
The dataset was made up of $239$ portal phase 2-D CT scans: $66$ cysts, $81$ metastases, $65$ hemangiomas, $27$ healthy. An expert radiologist marked the margin of each lesion and determined its corresponding diagnosis which was established by biopsy or a clinical follow-up. This serves as our ground truth. Fig. \ref{fig:data} shows a set of  data samples from the different categories. Perceptual similarities often exist across the class samples.  

\subsection{Evaluation}
We used a 3-fold cross validation to evaluate the classification performance with and without our proposed training process. In each fold 30\% of the training set was used as a validation set during the training process. Online data augmentation was performed showing each case $10$ times in each epoch with different affine transformations including scale, rotation, and translation. The results are presented in Table \ref{tab:acc}. Using our training scheme we achieved an improvement of $7.4\%$ on average with a similar trend in each fold.

Table \ref{tab:classic_conf_mat} 
shows the confusion matrices using the classic and proposed training scheme. Using the proposed training scheme the network was able to improve the overall classification performance but had more mistakes distinguishing between metastasis and hemangioma cases. These cases can be very similar as can be seen in Fig. \ref{fig:data} and additional work should be conducted to tackle this problem.

\begin{table}
	\centering
	\ra{1.3}
	\begin{tabu}{@{}clclc@{}}
    \toprule
		\textbf{Fold} & \phantom{ab} &  \textbf{Baseline [Acc.\%]} & \phantom{ab} & \textbf{Proposed [Acc.\%]}\\
    \midrule
		1 && 64.8  && 67.8  \\
		2 && 69.5  && 75.6  \\
		3 && 68.6  && 81.9  \\
        Avg. && 67.6 $\pm$ 2.0  && \textbf{75.0} $\pm$ 5.8  \\
	\bottomrule
	\end{tabu}
    \vspace{-0.3cm}
\caption{Accuracy in classification comparison with and without the proposed training scheme.}
\label{tab:acc}
\end{table}

\begin{table}
	\centering
	\ra{1.3}
	\begin{tabu}{c|cccc}
    \multicolumn{5}{c}{\emph{\textbf{Classic} training scheme}}   \\
    \midrule
\textbf{GT\textbackslash{}Predicted} & \textbf{Cyst} & \textbf{Met.} & \textbf{Hem.} & \textbf{Healthy} \\
    \midrule
\textbf{Cyst}                        & 60   & 5          & 1          & 0       \\
\textbf{Metastasis}                  & 13    & 43         & 19         & 6       \\
\textbf{Hemangioma}                  & 3    & 19         & 34         & 9       \\
\textbf{Healthy}                     & 0    & 0          & 2          & 25     \\
\bottomrule
\multicolumn{5}{c}{} \\
\end{tabu}
\begin{tabu}{c|cccc}
    \multicolumn{5}{c}{\emph{\textbf{Proposed} training scheme}}   \\
    \midrule
\textbf{GT\textbackslash{}Predicted} & \textbf{Cyst} & \textbf{Met.} & \textbf{Hem.} & \textbf{Healthy} \\
    \midrule
\textbf{Cyst}                        & \textbf{62}   & 3          & 1          & 0       \\
\textbf{Metastasis}                  & 5    & \textbf{50}         & 23         & 3       \\
\textbf{Hemangioma}                  & 3    & 20         & \textbf{41}         & 1       \\
\textbf{Healthy}                     & 0    & 0          & 0          & \textbf{27}     \\
\bottomrule
\end{tabu}
\vspace{-0.3cm}
\caption{\textbf{Classification confusion matrix:} \textit{\textbf{Top}:} using the \emph{classic} training scheme (i.e. the baseline). and \textit{\textbf{Bottom}:} using the \emph{proposed} training scheme. The proposed method improve the accuracy along all classes. }
\label{tab:classic_conf_mat}
\end{table}

\section{Conclusions}
  In this study we showed that using disentanglement of factors we can synthesize new mixtures of samples from different classes. Using the proposed classification target vector, we showed  
  superior accuracy for the liver lesion classification task. 
Future work includes comparison to other synthetic data generation schemes, as well as extend to more complex data augmentation techniques. In addition, we plan to test the generalization and robustness of our proposed approach on different datasets as well as better understand the different disentanglement factors that contribute to the classification decision using retrieval or attention maps.
\\

\noindent \textbf{Acknowledgements:} 
We thank Prof. Michal Amitai and Dr. Eyal Klang from the abdominal imaging unit at the Sheba Medical Center for providing the data used in this study.

\bibliographystyle{IEEEbib} 
\bibliography{refs}

\end{document}